\documentclass[10pt,twocolumn,letterpaper]{article}

\usepackage{iccv}
\usepackage{times}
\usepackage{epsfig}
\usepackage{graphicx}
\usepackage{amsmath}
\usepackage{amssymb}
\usepackage{graphicx}
\usepackage{diagbox}
\usepackage{rotating}
\usepackage{multirow}
\usepackage[normalem]{ulem}
\usepackage{caption}
\usepackage{subcaption}
\usepackage{booktabs}
\usepackage{colortbl}
\usepackage{array}
\usepackage{xcolor}
\usepackage[accsupp]{axessibility}  

\usepackage[pagebackref=true,breaklinks=true,letterpaper=true,colorlinks,bookmarks=false]{hyperref}

\iccvfinalcopy 

\def\iccvPaperID{5954} 

\ificcvfinal\fi

\newcommand\blfootnote[1]{%
  \begingroup
  \renewcommand\thefootnote{}\footnote{#1}%
  \addtocounter{footnote}{-1}%
  \endgroup
}

\begin{document}

\title{Tuning Pre-trained Model via Moment Probing}

\author{Mingze Gao$^{1,2,\dag}$\quad Qilong Wang$^{1, *}$\quad Zhenyi Lin$^{1}$\quad Pengfei Zhu$^{1}$\quad Qinghua Hu$^{1}$\quad Jingbo Zhou$^{2,*}$ \\
$^{1}$Tianjin Key Lab of Machine Learning, College of Intelligence and Computing, Tianjin University, China\\ \qquad $^{2}$Business Intelligence Lab, Baidu Research, China\\
{\tt\small \{gaomingze, qlwang, linzhenyi, zhupengfei, huqinghua\}@tju.edu.cn, zhoujingbo@baidu.com}
}

\maketitle
\ificcvfinal\thispagestyle{empty}\fi

\begin{abstract}
Recently, efficient fine-tuning of large-scale pre-trained models has attracted increasing research interests, where linear probing (LP) as a fundamental module is involved in exploiting the final representations for task-dependent classification. However, most of the existing methods focus on how to effectively introduce a few of learnable parameters, and little work pays attention to the commonly used LP module. In this paper, we propose a novel Moment Probing (MP) method to further explore the potential of LP. Distinguished from LP which builds a linear classification head based on the mean of final features (e.g., word tokens for ViT) or classification tokens, our MP performs a linear classifier on feature distribution, which provides the stronger representation ability by exploiting richer statistical information inherent in features. Specifically, we represent feature distribution by its characteristic function, which is efficiently approximated by using first- and second-order moments of features. Furthermore, we propose a multi-head convolutional cross-covariance (MHC$^3$) to compute second-order moments in an efficient and effective manner. By considering that MP could affect feature learning, we introduce a partially shared module to learn two recalibrating parameters (PSRP) for backbones based on MP, namely MP$_{+}$. Extensive experiments on ten benchmarks using various models show that our MP significantly outperforms LP and is competitive with counterparts at lower training cost, while our MP$_{+}$ achieves state-of-the-art performance. 
\end{abstract}
\blfootnote {$^\dag$ This work was done when Mingze Gao was an intern at Baidu Research. $^*$ Corresponding authors}
\section{Introduction}
Benefiting from the emergence of huge-scale datasets~\cite{5206848, sun2017revisiting, schuhmann2021laion}, the rapid development of neural network architectures~\cite{dosovitskiy2021an, he2016deep, tolstikhin2021mlp, liu2021swin} and self-supervised learning~\cite{he2022masked, radford2021learning, caron2021emerging}, large-scale pre-trained models dependent on sufficient computational resources show the great potential of the transferability on downstream tasks~\cite{devlin-etal-2019-bert, chen2021empirical, radford2019language, Language_Models_2020}, where full fine-tuning as a basic method has achieved promising performance.  However, full fine-tuning suffers from a high computational cost and is easy overfitting on small-scale datasets~\cite{Lian_2022_SSF, jia2022vpt}. In contrast, a simpler and more efficient method is only to tune a linear classifier (i.e., linear probing~\cite{he2020momentum}). Compared to full fine-tuning, linear probing (LP) usually suffers from inferior performance. To address this, existing works make a lot of efforts on parameter-efficient strategies~\cite{houlsby2019parameter, ben-zaken-etal-2022-bitfit, hu2022lora, li-liang-2021-prefix, karimi2021compacter, jia2022vpt, Lian_2022_SSF, chen2022adaptformer}. Going beyond LP, they focus on introducing a few learnable parameters to recalibrate features from frozen pre-trained models for downstream tasks. These methods avoid tuning massive parameters and show better efficiency and effectiveness trade-off. 

\begin{table}
\centering
\scalebox{0.82}{
\begin{tabular}{l|c|c|c|c|c} 
\specialrule{.1em}{0em}{0em} 
Method & \begin{tabular}[c]{@{}c@{}}IN-1K \\(\%)\end{tabular} & \begin{tabular}[c]{@{}c@{}}NABirds \\(\%)\end{tabular} & \begin{tabular}[c]{@{}c@{}}Params.\\(M)\end{tabular} & \begin{tabular}[c]{@{}c@{}}Time \\(ms)\end{tabular} & \begin{tabular}[c]{@{}c@{}}Mem.\\(G)\end{tabular}  \\ 
\hline
Linear probing   & 82.04 &75.9 & 0.77 & \textcolor{magenta}{\textbf{60}} & \textcolor{magenta}{\textbf{3.23}} \\ 
MP (Ours) & \textcolor{cyan}{\textbf{83.15}} &\textcolor{cyan}{\textbf{84.9}} & 3.65 & \textcolor{cyan}{\textbf{72}} & \textcolor{cyan}{\textbf{3.34}} \\
\hline
VPT-Shallow~\cite{jia2022vpt} & 82.08 &78.8 & 0.92 & 115 & 11.39 \\
VPT-Deep~\cite{jia2022vpt}  & 82.45 &84.2 & 1.23 & 120 & 11.39 \\
AdaptFormer~\cite{chen2022adaptformer} & 83.01 &84.7 & 1.07 & 125 & 10.49 \\
SSF~\cite{Lian_2022_SSF} & 83.10 &85.7 & 0.97 & 187 & 13.78 \\ 
\hline
Full fine-tuning & 83.58 &82.7 & 86.57 & 157 & 11.92 \\
MP$_{+}$ (Ours) &\textcolor{magenta}{\textbf{83.62}} &\textcolor{magenta}{\textbf{86.1}} & 4.10 & 140 & 11.22 \\
\specialrule{.1em}{0em}{0em} 
\end{tabular}}
\caption{Comparison of various tuning methods for pre-trained models in terms of recognition accuracy (\%), learnable parameters (Params.), training time (Time) per mini-batch, and GPU memory (Mem.) usage of training on ImageNet-1K (IN-1K) and NABirds, where ViT-B/16 pre-trained on IN-21K is used as basic backbone.}
\label{table 1}
\end{table}
 \begin{figure*}[t]
\begin{subfigure}{.62\textwidth}
\centering
\includegraphics[width=1.0\linewidth,trim=15 5 15 0,clip]{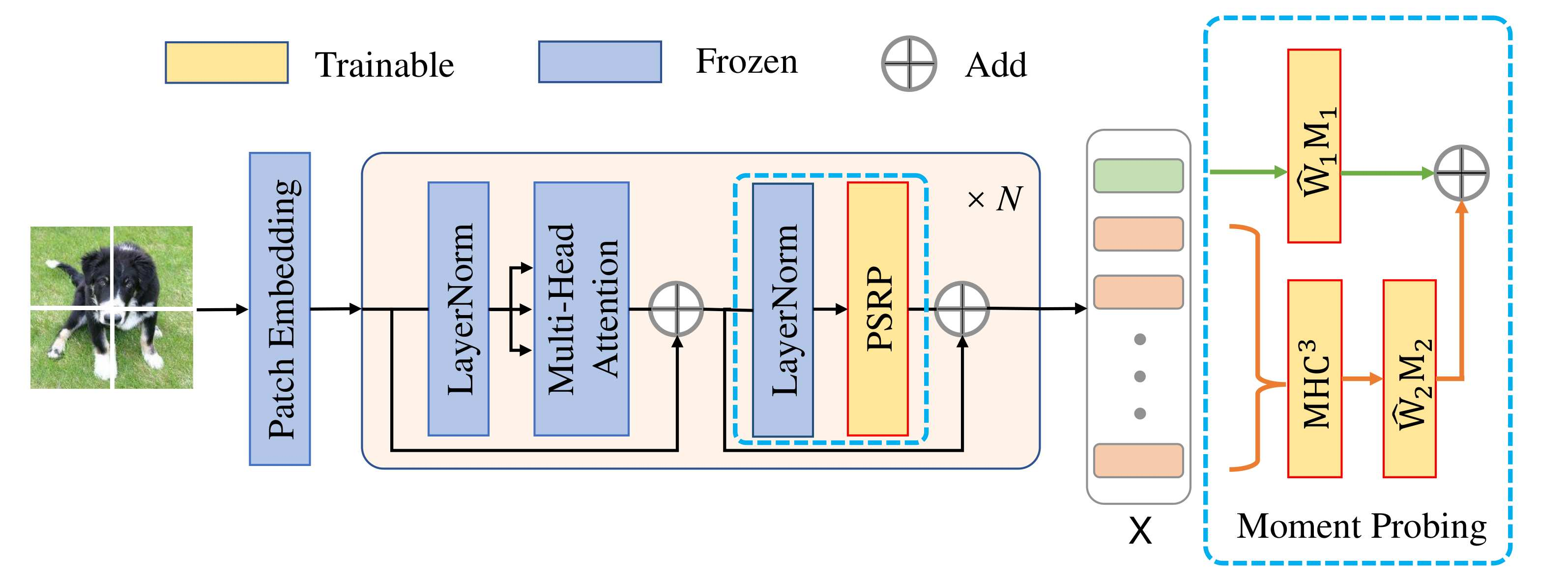}
\caption{}
\label{Architecture}
\end{subfigure}
\begin{subfigure}{.37\textwidth}
\centering
\includegraphics[width=1.0\linewidth,trim=0 5 0 5,clip]{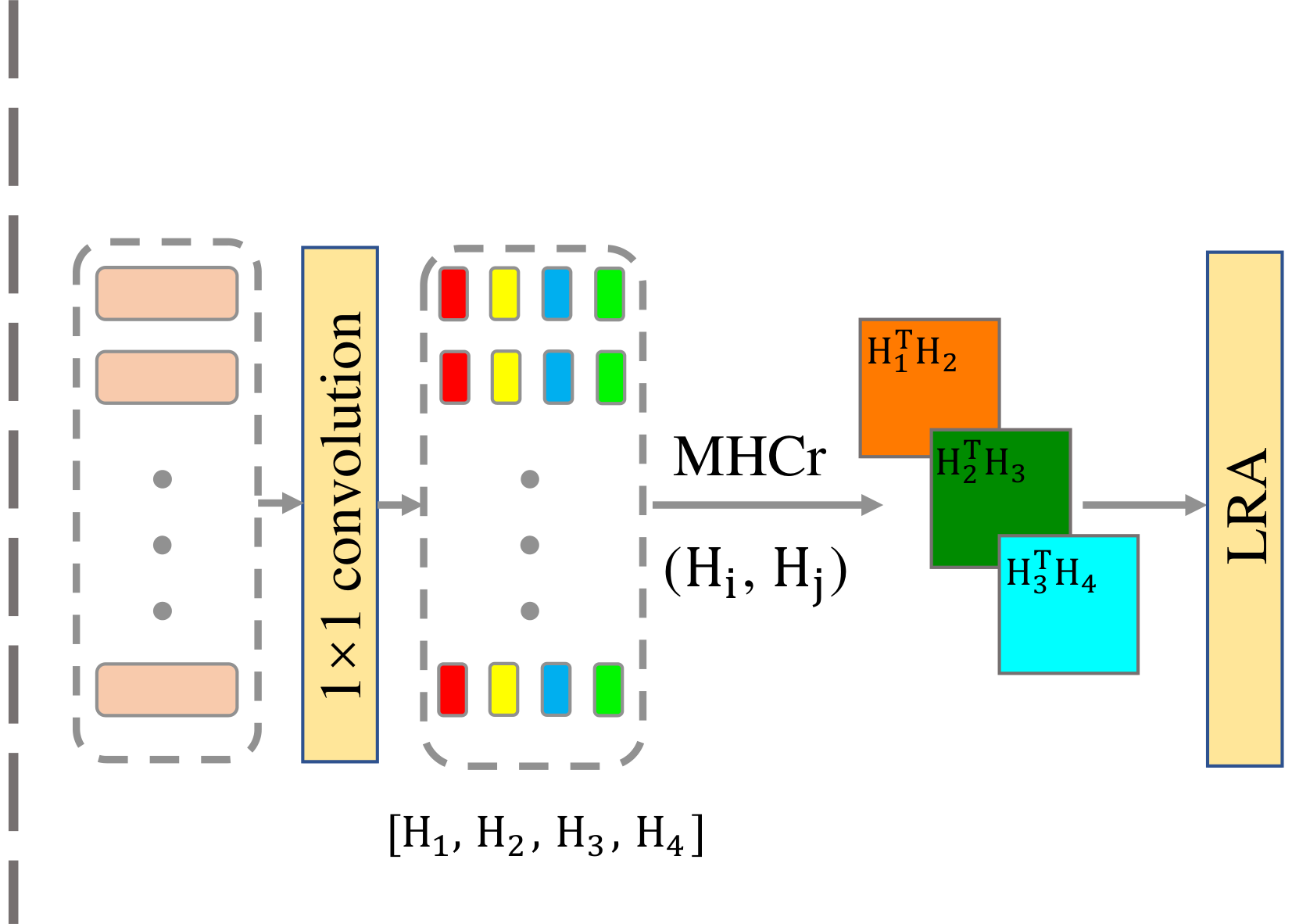}
\caption{}
\label{Adapt}
\end{subfigure}
\caption{(a) Overview of proposed MP$_+$ method for tuning pre-trained models, whose core is Moment Probing (MP) indicated by \textcolor[RGB]{0, 176, 240}{blue} dashed line instead of the original linear probing. Specifically, our MP performs a linear classifier on powerful representations characterized by feature distribution, which is approximated by using first- and second-order moments of features. To efficiently explore second-order moments, we present a (b) multi-head convolutional cross-covariance (MHC$^3$) method, whose details can refer to Sec.~\ref{sec:MP}. Besides, a partially shared module to learn two recalibrating parameters (PSRP) is introduced for exploring the potential of MP on feature learning.}
\label{Framework 1}
\end{figure*}

Although many advanced efforts are made on parameter-efficient tuning, little work pays attention to the most fundamental LP, which is involved in all existing tuning methods to learn a task-dependent classification head, and is closely related to the performance of downstream tasks. It can be observed that LP learns a linear classifier on input features, which are generally presented by classification token~\cite{dosovitskiy2021an}, average pooling (mean) of word tokens~\cite{tolstikhin2021mlp, liu2021swin} or convolution features~\cite{liu2022convnet}. From the statistical perspective, LP mainly exploits first-order statistics of features, taking no full merit of rich statistical information inherent in features.

In this paper, we further explore the potential of LP by generating more powerful representations for linear classifier. To this end, we propose a Moment Probing (MP) method, whose core is to perform a linear classifier on feature distribution, which portrays the full picture of features and provides more powerful representations. Specifically, we model probability density of features by their characteristic function, and approximate it by using first- and second-order moments of features for computational efficiency. Furthermore, we present a multi-head convolutional cross-covariance (MHC$^3$) method to efficiently compute second-order moments, avoiding the issues of computation and generalization brought by high-dimensional second-order representations. For computing MHC$^3$, we first split features into several groups and  compute cross-covariance between each two adjacent groups to extract second-order statistics. Then, a parameter-efficient module based on convolutions is designed to strengthen the interactions among cross-covariances and reduce the size of second-order representations. Compared to the original second-order moments, our MHC$^3$ performs better in terms of both efficiency and effectiveness. Table~\ref{table 1} shows our MP shares similar training cost with LP, but obtains much higher accuracy. Meanwhile, MP is comparable to or better than existing parameter-efficient methods at lower training cost.

Since deep models are trained in an end-to-end learning manner, the classifier will bring effect on feature learning. To explore the potential of our MP on feature learning, we introduce a partially shared module to learn two recalibrating parameters (PSRP) for pre-trained models, inspired by parameter-efficient methods~\cite{chen2022adaptformer,Lian_2022_SSF}. By combining MP with PSRP, our MP$_+$ surpasses full fine-tuning while learning much fewer parameters, whose overview is illustrated in Figure~\ref{Framework 1}. The contributions of our work are summarized as follows: (1) To our best knowledge, we make the first attempt to explore the potential of LP for tuning pre-trained models. To this end, we propose a Moment Probing (MP) method, which performs a linear classifier on powerful representations characterized by feature distribution. (2) For the efficiency of MP, we approximate feature distribution by using first- and second-order moments of features, and then present a multi-head convolutional cross-covariance (MHC$^3$) method to explore second-order moments in an efficient and effective manner. (3) By considering the effect of our MP on feature learning, we introduce a partially shared module to learn two recalibrating parameters (PSRP), resulting in a MP$_+$ method. It further exploits the potential of our MP in tuning  pre-trained models. (4) We conduct extensive experiments on ten benchmarks using various models, and results show our MP is superior to LP while generalizing well to pre-training strategies, few-shot and out-of-distribution settings. Besides, our MP$_{+}$ outperforms existing parameter-efficient methods, while achieving state-of-the-art performance.

\section{Related Work}
\subsection{Transfer Learning}
 Transfer learning aims to reuse the pre-trained models and re-adapting them to new tasks by fine-tuning them on a specific dataset. In both of natural language processing and computer vision communities, transferring pre-trained large models to downstream tasks has long been a popular paradigm~\cite{devlin-etal-2019-bert, ben-zaken-etal-2022-bitfit,zhuang-etal-2021-robustly, Language_Models_2020, he2022masked, radford2021learning, caron2021emerging, chen2021empirical}, which always provides rewarding performance with the total amount of pre-trained data and the scale of the model itself increase. During the transfer process, tuning strategies are limited to basic full fine-tuning and linear probing, suffering from low parameter efficiency and inferior performance. To solve above problem, recent works attempt to explore parameter-efficient methods by tuning a few of learnable parameters in the frozen backbone and sharing most of the parameters in the backbone for each downstream task~\cite{hu2022lora, houlsby2019parameter, ben-zaken-etal-2022-bitfit,Lian_2022_SSF, jia2022vpt, chen2022adaptformer}. For example, some works selectively fine-tune the original parameters in the backbone where SpotTune~\cite{guo2019spottune} investigates which layers need to tune and Bitfit~\cite{ben-zaken-etal-2022-bitfit} finds that fine-tuning bias term is enough in some cases. Other works focus on inserting additional modules to adapt feature transfer. Among them, Adaptor-based methods~\cite{houlsby2019parameter, karimi2021compacter} insert an additional non-linear MLP block between frozen layers. VPT~\cite{jia2022vpt} adds learnable prompt tokens into the input space of frozen backbone. SSF~\cite{Lian_2022_SSF} introduces both the scale and shift factors after each operation in backbone. AdaptFormer~\cite{chen2022adaptformer} proposes a parallel branch adding to the feed-forward network (FFN) of the pre-trained model. Different from aforementioned works, our MP makes the first attempt to explore the potential of LP, while achieving comparable or better performance than above counterparts at much lower training cost. 

\subsection{Second-order Pooling}
Second-order moment involved in our MP is closely related to global covariance pooling (GCP)~\cite{lin2015bilinear}, which has been studied to improve the representation and generalization abilities of deep architectures. Previous works~\cite{lin2015bilinear, BMVC2017_117, li2018towards, li2017second} have shown that GCP is an effective alternative to global average pooling (GAP) for visual tasks. For example, B-CNN~\cite{lin2015bilinear} and MPN-COV~\cite{li2018towards} insert GCP with different post-normalization methods into deep convolutional neural networks (CNNs) for fine-grained and large-scale visual, respectively. Recently, DropCov~\cite{wangdropcov} proposes an adaptive channel dropout method for GCP normalization, and shows the effectiveness in both deep CNNs and ViTs. Besides, some researches focus on reducing the dimension of GCP representations, resulting in several compact models~\cite{gao2016compact, kong2017low}. Different from the above GCP methods, our MP presents a multi-head convolutional cross-covariance (MHC$^3$) to efficiently compute the second-order moment, which is designed for parameter-efficient tuning of pre-trained models and outperforms existing counterparts in terms of efficiency and effectiveness.

 \section{Proposed Method}
In this section, we will introduce the proposed MP$_{+}$ method. As shown in Figure~\ref{Framework 1}, the core of MP$_+$ is a Moment Probing (MP) instead of the original LP for tuning pre-trained models.  Besides, a partially shared module to learn two recalibrating parameters (PSRP) is introduced to explore the potential of MP for feature learning. In the following, we will describe our MP and PSRP in detail.
\subsection{Moment Probing}\label{sec:MP}
\noindent \textbf{Linear Classifier on Feature Distribution.} Given a set of $N$ $d$-dimensional features $\mathbf{X}\in \mathbb{R}^{N\times d}$ output from the block right before classifier of pre-trained models (e.g., final word tokens for ViTs and last convolution features for CNNs), linear probing (LP) builds a linear classifier on a representation $g(\mathbf{X})$ generated by $\mathbf{X}$, i.e.,  
\begin{equation}
\mathbf{y}_{pred} = \mathbf{W}g(\mathbf{X}),
\label{formula 1}
\end{equation}
where $\mathbf{W}\in \mathbb{R}^{C \times S}$ indicates weights of classifier. $C$ and $S$ are class number of downstream task and the size of representation $g(\mathbf{X})$, respectively. For the original LP, $g(\mathbf{X})$ is usually generated by classification token~\cite{dosovitskiy2021an} (weighted combination of word tokens $\mathbf{X}$), average (mean) pooling  of $\mathbf{X}$ (e.g., word tokens ~\cite{tolstikhin2021mlp, liu2021swin} or convolution features~\cite{liu2022convnet}).  From the statistical perspective, $g(\mathbf{X})$ of LP mainly exploits first-order statistics of features $\mathbf{X}$. As shown in Eqn.~(\ref{formula 1}), prediction $\mathbf{y}_{pred}$ is definitely influenced by representation $g(\mathbf{X})$. Therefore, stronger representations potentially make the prediction more precise. 

To this end, we propose a Moment Probing (MP) to generate stronger representations by exploiting probability distribution $P(\mathbf{X})$ of features $\mathbf{X}$ instead of simple mean point in $g(\mathbf{X})$, because $P(\mathbf{X})$ can characterize the full picture of $\mathbf{X}$. As such, our MP aims to perform a linear classifier on feature distribution $P(\mathbf{X})$: 
\begin{equation}
\mathbf{y}_{pred} = \widehat{\mathbf{W}}P(\mathbf{X}).
\label{formula-1-1}
\end{equation}
However, $P(\mathbf{X})$ is usually unknown. Based on the classical probability theory~\cite{bishop2006pattern}, $P(\mathbf{X})$ can be defined by its characteristic function $\varphi_{\mathbf{X}}(t)$, according to Fourier transforms between $\varphi_{\mathbf{X}}(t)$ and probability density function $p(\mathbf{X})$ as 
\begin{equation}
 \varphi_{\mathbf{X}}(t) = E\left[e^{it\mathbf{X}}\right] = \int_{\mathbb{R}}e^{it\mathbf{x}}dP_{\mathbf{X}}(\mathbf{x}) = \int_{\mathbb{R}}e^{it\mathbf{x}}p_{\mathbf{X}}(\mathbf{x})d\mathbf{x},
\label{formula 2}
\end{equation}
 where $i$ is the imaginary unit, and $t \in \mathbb{R}$ is the argument of the characteristic function. Furthermore, we can rewrite $\varphi_{\mathbf{X}}(t)$ with the Taylor series of $e^{it\mathbf{X}}$ as: 
\begin{equation}
\begin{aligned}
  \varphi_{\mathbf{X}}(t) &= E\left[\sum_{k=0}^{\infty}\frac{(it\mathbf{X})^k}{k!}\right] = \sum_{k=0}^{\infty}\frac{(it)^k}{k!}E\left[\mathbf{X}^{k}\right]\\
 &=1 + (it)E\left[\mathbf{X}\right] + \frac{(it)^{2}}{2!}E\left[\mathbf{X}^{2}\right] + \cdots,
\end{aligned}
\label{formula 3}
 \end{equation}
where $E\left[\mathbf{X}^{k}\right]$ indicates $k^{\text{th}}$-order moment ($\mathbf{M}_{k}$) of $\mathbf{X}$. Let the coefficient of $\mathbf{M}_{k}$ be $\omega_k$, we can rewrite Eqn.~(\ref{formula 3}) as 
 \begin{equation}
\varphi_{\mathbf{X}}(t) = 1 + {\omega}_1\mathbf{M}_1 + {\omega}_2{\mathbf{M}}_2 + \cdots = 1 + \sum_{k=1}^{\infty}{\omega}_k\mathbf{M}_k,
\label{formula 4}
\end{equation}
 
So far, we can represent feature distribution $P(\mathbf{X})$ by using its characteristic function as for Eqn.~(\ref{formula 4}). By omitting the constant term, we reformulate Eqn.~(\ref{formula-1-1}) as:
\begin{equation}
\mathbf{y}_{pred} =\widehat{\mathbf{W}}\left(\sum_{k=1}^{\infty}{\omega}_k\mathbf{M}_k\right) = \sum_{k=1}^{\infty}\left(\widehat{\mathbf{W}}_k\mathbf{M}_k\right),
\label{formula 5}
\end{equation}
where we can see that $\mathbf{y}_{pred}$ in Eqn.~(\ref{formula 5}) fuses prediction results from different order moments, which are superior to single first-order statistics in the original LP. Although combination of more statistics generally leads to a more precise approximation on feature distribution, it brings much more computational cost and is sensitive to noise~\cite{zhang2022modern}. Therefore, our MP exploits the first- and second-order moments to approximate feature distribution for final prediction:
\begin{equation}
\mathbf{y}_{pred} = \sum_{k=1}^2\left(\widehat{\mathbf{W}}_k\mathbf{M}_k\right) = \widehat{\mathbf{W}}_1\mathbf{M}_1 + \widehat{\mathbf{W}}_2\mathbf{M}_2,
\label{formula 6}
\end{equation}
where $\mathbf{M}_1$ can be computed as classification token or average pooling of word tokens as for LP, while $\mathbf{M}_2$ is calculated by $\mathbf{X}^{T}\mathbf{X}$.
\begin{figure}[t]
\centering
   \includegraphics[width=1.0\linewidth,trim=0 15 0 15,clip]{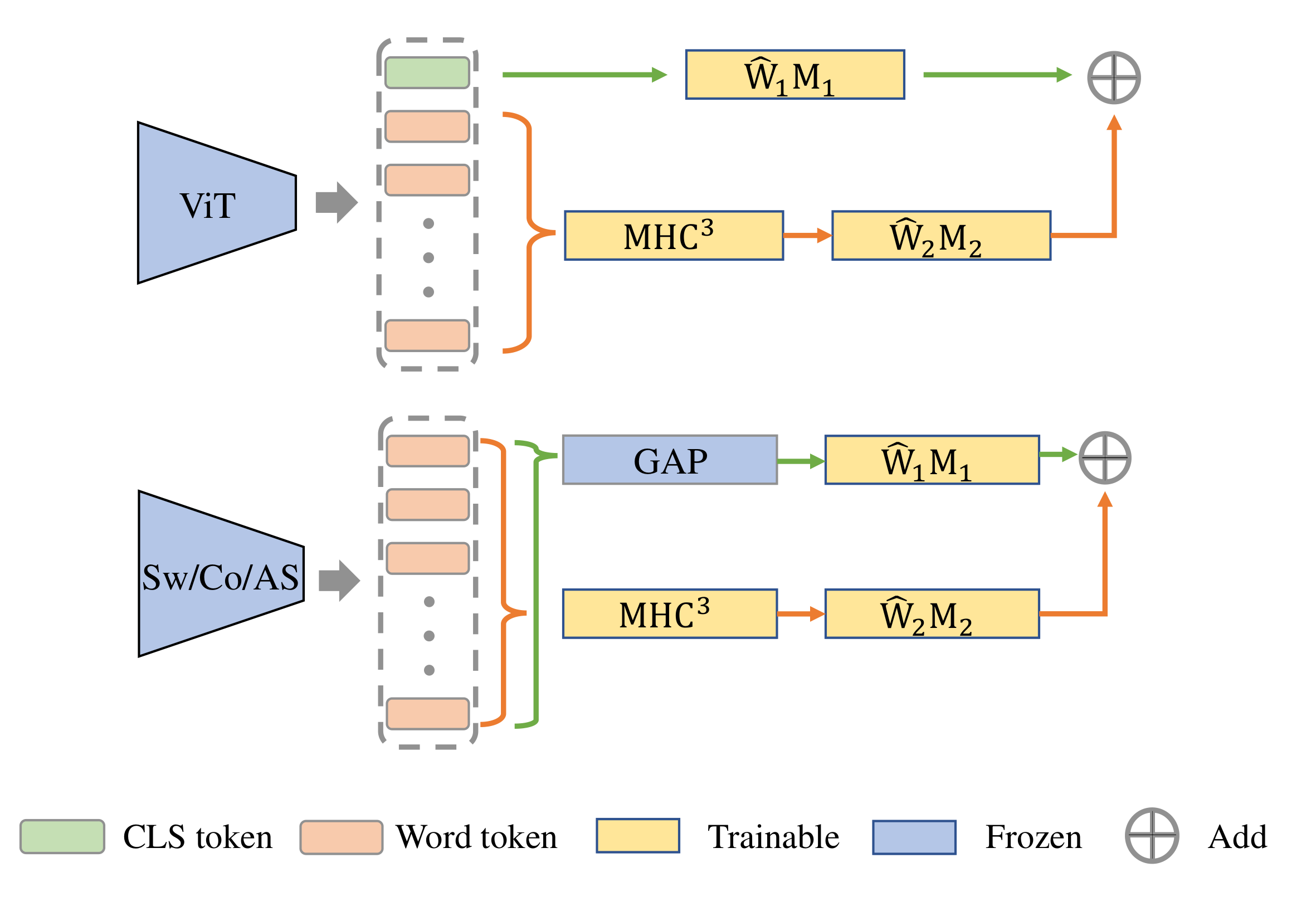}
   \caption{Usage details of MP for ViT, Swin transformer (Sw), ConvNeXt (Co) and AS-MLP (AS).}
\label{differ_usage}
\end{figure}
\\[5pt]
\noindent \textbf{Efficient Second-order Moment.} For the input features $\mathbf{X}\in \mathbb{R}^{N\times d}$, their second-order moment ($\mathbf{M}_2=\mathbf{X}^{T}\mathbf{X}$) results in a $d^2$-dimensional representation. As such, $\mathbf{M}_2$ will create large-size representations for high-dimensional features, bringing the issues of computational burden and overfitting. To handle above issues, we present a multi-head convolutional cross-covariance (MHC$^3$) to efficiently compute the second-order moment. As shown in Figure~\ref{Adapt}, we first exploit a $1\times1$ convolution to reduce dimension of $\mathbf{X}$ from $d$ to $\hat{d}$ ($\hat{d}<d$), which are indicated by $\widehat{\mathbf{X}}$. Then, features $\widehat{\mathbf{X}}$ are split into $h$ heads along feature dimension: 
\begin{equation}
 [\mathbf{H}_1; \mathbf{H}_2; \cdots; \mathbf{H}_h] = {\rm SP}\big(\widehat{\mathbf{X}}\big),
\label{formula 7}
\end{equation}
where $\rm SP$ indicates a splitting operation, and $\mathbf{H}_i\in \mathbb{R}^{N\times (\hat{d}/h)}$ are split features. To capture second-order statistics, we compute mutli-head cross-covariance between two adjacent split features, i.e.,
\begin{equation}
\mathbf{Z}_i= {\rm MHCr}(\mathbf{H}_i, \mathbf{H}_{(i+1)}) = \ell_{2}\left(\mathbf{H}_i^T\mathbf{H}_{(i+1)}\right),
\label{formula 8}
\end{equation}
where $\mathbf{Z}_i$ leads to a $(\hat{d}/h)^2$-dimensional cross-covariance representation, which captures second-order statistics between $\mathbf{H}_i$ and $ \mathbf{H}_{(i+1)}$. $\ell_{2}$ is an element-wise $\ell_{2}$ normalization to control the magnitude of representation $\mathbf{Z}_i$. 

It can be seen that $\mathbf{Z}=\{\mathbf{Z}_1; \mathbf{Z}_2; \cdots; \mathbf{Z}_{h-1}\}$ only capture second-order statistics between adjacent split features. To perform interaction among all features and further reduce representation size, we introduce a parameter-efficient Local Representation Aggregation (LRA) block for $\mathbf{Z}$:
\begin{equation}
\rm LRA(\mathbf{Z})=\rm Concat[\rm Conv_{3\times3}^{2}(\sigma(\rm Conv_{3\times3}^{2}(\mathbf{Z})))],
\label{formula 9}
\end{equation}
where $\rm Conv_{3\times3}^{2}$ indicates a $3\times3$ convolution with the stride of 2, while input channel and output one of convolution are $h-1$. $\sigma$ is GELU non-linearity, and $\rm Concat$ is a concatenated operation. By using Eqn.~(\ref{formula 7})$\sim$Eqn.~(\ref{formula 9}), we can compute the proposed ${\rm MHC^3}(\mathbf{X})$, which results in a  $(h-1)(\hat{d}/4h)^2$-dimensional cross-covariance representation. Compared with $d^2$-dimensional $\mathbf{M}_2=\mathbf{X}^{T}\mathbf{X}$, size of our ${\rm MHC^3}(\mathbf{X})$ is at least $16h$ times less than the original second-order moment. Based on the proposed MHC$^3$,  our MP is finally computed by
\begin{equation}
\mathbf{y}_{pred} = \widehat{\mathbf{W}}_1\mathbf{M}_1 + \widehat{\mathbf{W}}_2 \rm MHC^3(\mathbf{X}).
\label{formula 10}
\end{equation}  

\begin{figure}[t]
\centering
   \includegraphics[width=1.0\linewidth,trim=0 0 0 0,clip]{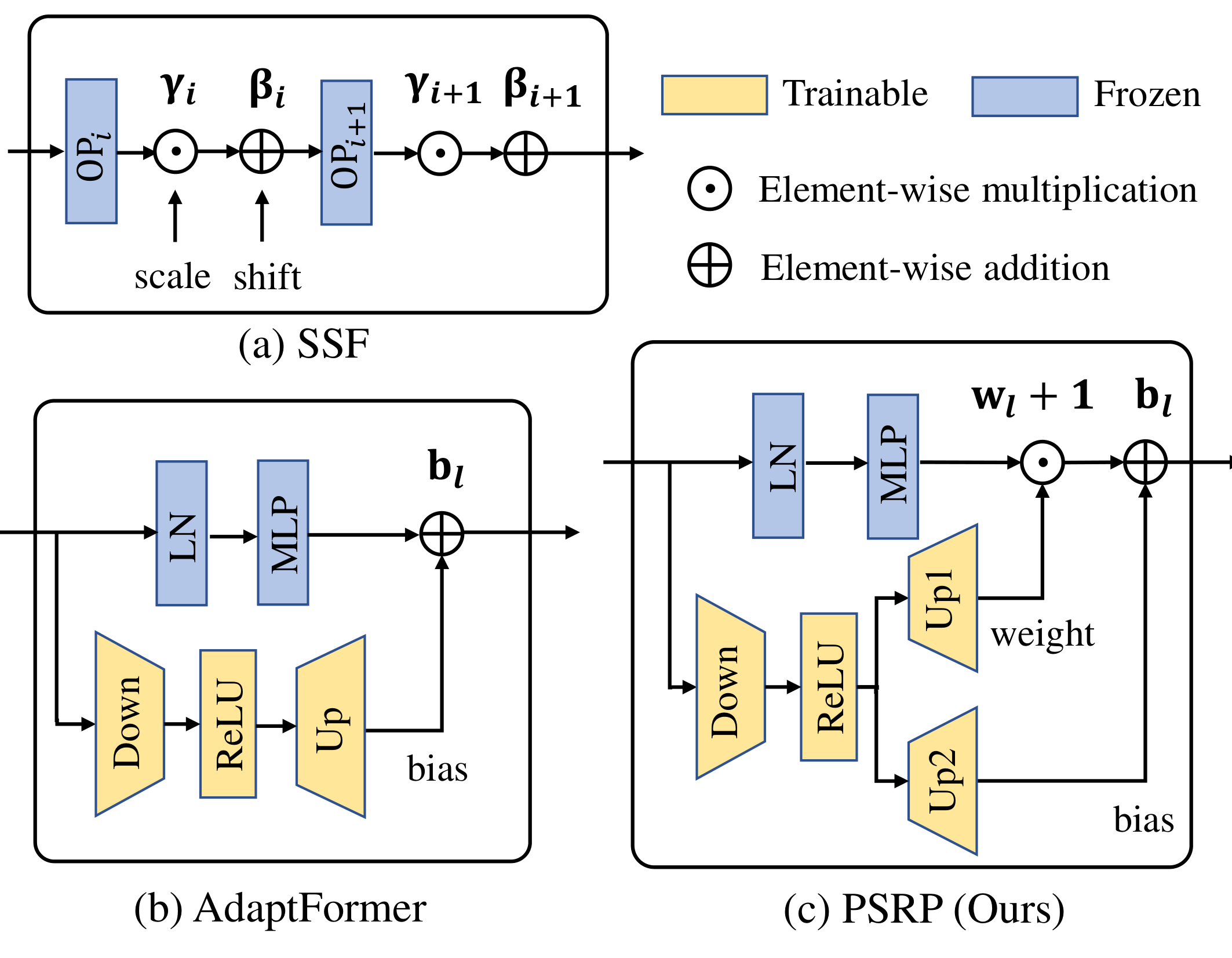}
   \caption{Comparison of (c) our PSRP with (a) SSF and (b) AdaptFormer, where SSF respectively introduces parameters $\boldsymbol\gamma_{i}$ and $\boldsymbol\beta_{i}$ for scaling and shifting features after each operation OP$_{i}$. AdaptFormer designs a tiny network to learn shifting parameters $\mathbf{b}_{i}$ for feature recalibration. Differently, our PSRP develops a partially shared module to learn both scaling and shifting parameters.}
\label{Framework 2}
\end{figure}

\noindent \textbf{Usage Details of MP.} We provide details on how to apply our MP for different model families. Specifically, as shown in Figure~\ref{differ_usage},  for ViTs~\cite{dosovitskiy2021an} we utilize the classification token (CLS\_token) as the first-order moment and compute the second-order moment by using MHC$^3$ of the final word tokens, which constitutes our MP. For Swin transformer (Swin)~\cite{liu2021swin}, ConvNeXt~\cite{liu2022convnet} and AS-MLP~\cite{Lian_2021_ASMLP}, we perform global average pooling (GAP) of the final word tokens (i.e., Swin/AS-MLP) or the final convolution features (ConvNeXt) as the first-order moment, while these features are feed to MHC$^3$ for computing the second-order moment. Finally, predication scores of first- and second-order moments are summed for fusion. 

\subsection{PSRP for Feature Recalibration}
To further explore the potential of MP, we investigate the effect of MP on learning intermediate features. To avoid tuning a number of parameters in large-scale pre-trained models, recently proposed parameter-efficient methods learn a few of additional parameters to recalibrate features from frozen backbones. Particularly, as shown in Figure~\ref{Framework 2} (a), SSF~\cite{Lian_2022_SSF} introduces learnable parameters $ \boldsymbol\gamma_{i} \in \mathbb{R}^{d_{i}}$ and $\boldsymbol\beta_{i} \in \mathbb{R}^{d_{i}}$ for scaling and shifting the output from each operation OP$_{i}(\mathbf{X}_{i})$ (e.g., multi-head self-attention or feed forward network (FFN) ) of pre-trained model:
\begin{equation}
\mathbf{Y}_{i} = \boldsymbol\gamma_{i} \odot {\rm OP}_{i}  (\mathbf{X}_{i}) \oplus \boldsymbol\beta_{i},
\label{formula 11}
\end{equation}  
where $d_{i}$ dimension $\mathbf{X}_{i}$ are inputs of $i$-operation OP$_{i}$ and $\mathbf{Y}_{i}$ are recalibrated features. $\odot$ and $\oplus$ indicate element-wise multiplication and addition along dimension, respectively. As for Eqn.~(\ref{formula 11}), the parameters $\boldsymbol\gamma_{i}$ and $\boldsymbol\beta_{i}$ are irrelevant to $\mathbf{X}_{i}$, neglecting effect of input. As shown in Figure~\ref{Framework 2} (b), AdaptFormer~\cite{chen2022adaptformer} designs an input-based tiny network to learn shifting parameters for feature recalibration in FFN, but it ignores the effect of scaling parameters. As suggested in SSF~\cite{Lian_2022_SSF}, both scaling and shifting operations help model transfer in downstream tasks.  

Based on above analysis, we introduce two input-based tiny networks to learn both scaling and shifting parameters for feature recalibration. By considering parameter efficiency, we develop a partially shared module to learn two recalibrating parameters (PSRP). As shown in Figure~\ref{Framework 2} (c), our PSRP learns scaling ($\mathbf{w}_{l}$) and shifting ($\mathbf{b}_{l}$) parameters of $l^{\text{th}}$-layer outputs as
\begin{equation}
\begin{split}
{\mathbf{w}_{l}} = {\mathbf{W}}_{up}^{1}({\rm ReLU}({\mathbf{W}}_{down}(\mathbf{X}_{l}))),\\
{\mathbf{b}_{l}} = {\mathbf{W}}_{up}^{2}({\rm ReLU}({\mathbf{W}}_{down}(\mathbf{X}_{l}))),
\end{split}
\label{formula 12}
\end{equation}
where $\mathbf{X}_{l} \in \mathbb{R}^{d_{l}}$ are inputs of $l^{\text{th}}$-layer, while $\mathbf{W}_{up}^{1} \in \mathbb{R}^{d_{l} \times d_{h}}$, $\mathbf{W}_{up}^{2} \in \mathbb{R}^{d_{l} \times d_{h}}$ and $\mathbf{W}_{down} \in \mathbb{R}^{d_{h} \times d_{l}}$ are learnable parameters with hidden dimension of $d_{h}$. Finally, we obtain the recalibrated features by using $\mathbf{w}_{l}$ and $\mathbf{b}_{l}$ as  
\begin{equation}
\mathbf{Y}_{l} = (\mathbf{w}_{l}+\mathbf{1}) \odot{\rm FFN(LN}(\mathbf{X}_{l})) \oplus \mathbf{b}_{l},
\label{formula 13}
\end{equation}
where $\rm LN$ indicates layer normalization~\cite{ba2016layer}, and $\mathbf{1}$ is a vector of all ones.

\begin{table*}[t]
\centering
\begin{tabular}{l|c|c|c|c|c|c|c|c} 
\specialrule{.1em}{0em}{0em} 
\diagbox{Method}{Dataset} & \multicolumn{1}{l|}{CIFAR-100} & \begin{tabular}[c]{@{}c@{}}CUB-200\\-2011\end{tabular} & \multicolumn{1}{l|}{NABirds} & \begin{tabular}[c]{@{}c@{}}Oxford\\Flowers\end{tabular} & \begin{tabular}[c]{@{}c@{}}Stanford\\Dogs\end{tabular} & \begin{tabular}[c]{@{}c@{}}Stanford\\Cars\end{tabular} & Mean  & \begin{tabular}[c]{@{}c@{}}Params.\\(M)\end{tabular}           \\ 
\hline
Linear probing & 88.7 & 85.3 & 75.9 & 97.9 & 86.2 & 51.3 & 80.88 & 0.17\\ 
MP~(Ours) & \textbf{\textcolor{blue}{93.8$_{(5.1)}$}} & 89.3$_{(4.0)}$ & 84.9$_{(9.0)}$ & \textbf{\textcolor{blue}{99.6$_{(1.7)}$}} & 89.5$_{(3.3)}$ & 83.6$_{(32.3)}$ & 90.12$_{(9.24)}$ &1.20\\
\hline
Adapter \cite{houlsby2019parameter} & 93.3 & 87.1 & 84.3 & 98.5 & 89.8 & 68.6 & 86.93 & 0.40\\
Bias~\cite{ben-zaken-etal-2022-bitfit} & 93.4 & 88.4 & 84.2 & 98.8 & \textbf{\textcolor{red}{91.2}} & 79.4 & 89.23 & 0.27\\
VPT-Shallow~\cite{jia2022vpt} & 90.4 & 86.7 & 78.8 & 98.4 & \textbf{\textcolor{blue}{90.7}} & 68.7 & 85.62 & 0.25\\
VPT-Deep \cite{jia2022vpt} & 93.2 & 88.5 & 84.2 & 99.0 & 90.2 & 83.6 & 89.78 & 0.81\\
AdaptFormer~\cite{chen2022adaptformer} &93.6 & 88.4 &84.7 &99.2 &88.2 &81.9&89.33 & 0.46\\
SSF~\cite{Lian_2022_SSF} & 94.0 & \textbf{\textcolor{blue}{89.5}} & \textbf{\textcolor{blue}{85.7}} & \textbf{\textcolor{blue}{99.6}} & 89.6 & \textbf{\textcolor{blue}{89.2}} & \textbf{\textcolor{blue}{91.27}} & 0.38 \\ 
\hline
Full fine-tuning & 93.8 & 87.3 & 82.7 & 98.8 & 89.4 & 84.5 & 89.42 & 85.96 \\
MP$_{+}$~(Ours)& \textbf{\textcolor{red}{94.2}}&\textbf{\textcolor{red}{89.9}} &\textbf{\textcolor{red}{86.1}}& \textbf{\textcolor{red}{99.7}} &90.2 &\textbf{\textcolor{red}{89.4}} &\textbf{\textcolor{red}{91.58}} & 1.64\\
\specialrule{.1em}{0em}{0em} 
\end{tabular}
\caption{Comparison of various fine-tuning methods on different downstream tasks (i.e., CIFAR-100 and FGVC datasets), where ViT-B/16 model pre-trained on ImageNet-21K is used as basic backbone.}
\label{table 2}
\end{table*}

\section{Experiments}
In this section, we first describe the experimental settings, and then compare with state-of-the-art (SOTA) methods on various downstream datasets as well as using different pre-trained models. Sec.~\ref{4.3} conducts ablation studies to assess the effect of key components on our MP. Additionally, we evaluate generalization ablity of our MP to out-of-distribution datasets, other parameter-efficient methods, pre-training strategies and few-shot setting in Sec.~\ref{4.4}, Sec.~\ref{4.5} Sec.~\ref{4.6} and Sec.~\ref{4.7}, respectively.

\subsection{Experimental Settings}
\label{4.1}
\noindent \textbf{Pre-trained Backbone.} 
In our experiments, we adopt four kinds of backbone models, including ViT~\cite{dosovitskiy2021an}, Swin Transformer~\cite{liu2021swin}, ConvNeXt~\cite{liu2022convnet}, and AS-MLP~\cite{Lian_2021_ASMLP}. Specifically, we first employ the pre-trained ViT-B/16, Swin-B, ConvNeXt-B, and AS-MLP to compare with SOTA methods on downstream tasks. Then, ViT-B/16 pre-trained on ImageNet-21K is used to evaluate the generalization of our MP to out-of-distribution and few-shot settings. Additionally, we employ  ViT-Base/16, ViT-Large/14, ViT-Large/16 and ViT-Huge/14 for assessing generalization to different pre-training strategies (e.g., MAE~\cite{he2022masked} and CLIP~\cite{radford2021learning} ).
\\[5pt]
\noindent \textbf{Datasets.} Following the settings in SSF~\cite{Lian_2022_SSF}, we employ five datasets (i.e., CUB-200-2011~\cite{wah2011caltech}, NABirds~\cite{Horn_2015_CVPR}, Oxford Flowers~\cite{4756141}, Stanford Dogs~\cite{khosla2011novel}, and Stanford Cars~\cite{gebru2017fine}) for fine-grained visual classification. Besides, CIFAR-100~\cite{krizhevsky2009learning} and  ImageNet-1K~\cite{5206848} are used for general image classification. Additionally, ImageNet-A~\cite{hendrycks2021natural}, ImageNet-R~\cite{hendrycks2021many} and ImageNet-C~\cite{hendrycks2019robustness} are considered to evaluate the robustness of our method to out-of-distribution. 
\\[5pt]
\noindent \textbf{Implementation Details.} For fine-tuning   models pre-trained by fully-supervised scheme~\cite{dosovitskiy2021an} and CLIP~\cite{radford2021learning}, we follow the same settings in~\cite{dosovitskiy2021an, Lian_2022_SSF}. Specifically, for data augmentation, the input images are cropped to 224$\times$224 with a random horizontal flip for FGVC datasets, while stronger data augmentation strategies~\cite{dosovitskiy2021an} are adopted for CIFAR-100 and ImageNet-1K. AdamW~\cite{loshchilov2018decoupled} with warmup and cosine annealing schedule of learning rate is used for network optimization. For fine-tuning models pre-trained by MAE~\cite{he2022masked}, we follow its official configurations of LP on ImageNet-1K, which adopt a linear scaling rule~\cite{goyal2017accurate}. Refer to the supplementary materials for more details. Source code will be available at \href{https://github.com/mingzeG/Moment-Probing}{\textbf{https://github.com/mingzeG/Moment-Probing}}
\begin{table}[]
\centering
\scalebox{0.73}{
\begin{tabular}{l|c|c|c|c} 
\specialrule{.1em}{0em}{0em} 
\diagbox{Method}{Model} & ViT-B/16 & Swin-B & ConvNeXt-B & AS-MLP-B   \\ 
\hline
Linear probing & 82.04 & 83.25 & 84.05 & 79.04 \\
MP~(Ours) & 83.15$_{(1.11)}$ & 84.62$_{(1.37)}$  & 85.09$_{(1.04)}$ & 84.03$_{(4.99)}$ \\
\hline
Adapter~\cite{houlsby2019parameter}& 82.72 & 83.82 & 84.49 & 88.01 \\
Bias~\cite{ben-zaken-etal-2022-bitfit}& 82.74 & 83.92 & 84.63 & 87.46 \\
VPT-Shallow~\cite{jia2022vpt} & 82.08 & 83.29 & - & - \\
VPT-Deep~\cite{jia2022vpt}& 82.45 & 83.44  & - & - \\
AdaptFormer~\cite{chen2022adaptformer}& 83.01 & 84.08 & 84.79 & 88.86 \\
SSF~\cite{Lian_2022_SSF} & 83.10 & 84.40 & 84.85 & 88.28 \\ 
\hline
Full fine-tuning &\textbf{\textcolor{blue}{83.58}} &\textbf{\textcolor{red}{85.20}}  &\textbf{\textcolor{red}{85.80}} &\textbf{\textcolor{red}{89.96}} \\
MP$_{+}$~(Ours) & \textbf{\textcolor{red}{83.62}} & \textbf{\textcolor{blue}{84.95}}&\textbf{\textcolor{blue}{85.37}}  & \textbf{\textcolor{blue}{89.82}} \\
\specialrule{.1em}{0em}{0em} 
\end{tabular}}
\caption{Comparison of various fine-tuning methods using different backbones, where ViT-B/16, Swin-B, and ConvNeXt-B are pre-trained on ImageNet-21K and fine-tuned on ImageNet-1K. AS-MLP-B is pre-trained on ImageNet-1K and fine-tuned on CIFAR-100.}
\label{table 3}
\end{table}
\subsection{Comparison with SOTA}
\label{4.2}
\begin{figure}[t]
   \includegraphics[width=1.0\linewidth, trim=0 15 0 10,clip]{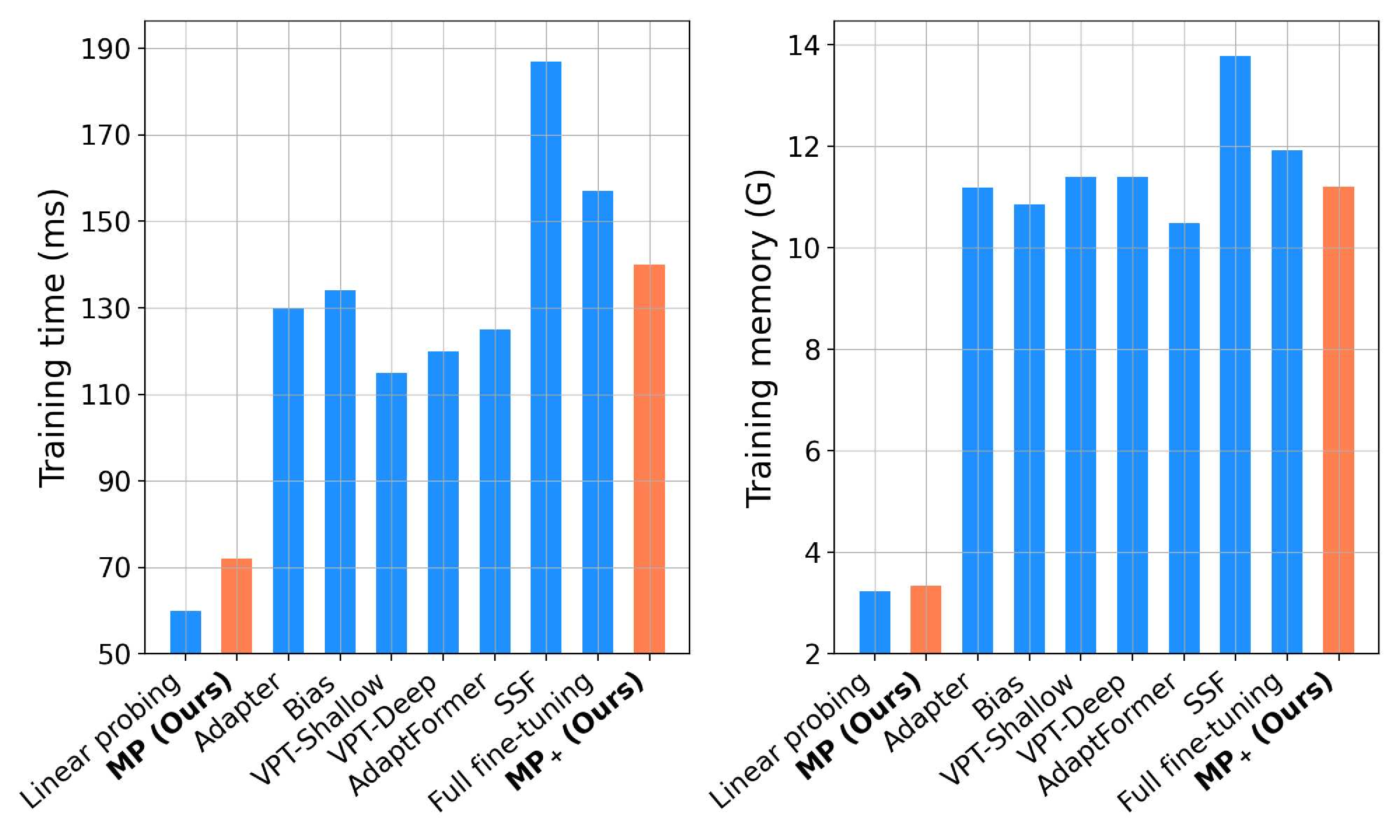}
\caption{Comparison of computational complexity for various fine-tuning methods in terms of training time and training memory.}
\label{cost}
\end{figure}

To verify the effectiveness of our MP, we compare with several SOTA fine-tuning methods on seven downstream classification tasks. Specifically, we compare with SOTA methods on CIFAR-100 and FGVC datasets, where ViT-B/16 model pre-trained on ImageNet-21K is used as a basic backbone. Besides, we compare with SOTA methods using various backbone models on ImageNet-1K and CIFAR-100. Top-1  accuracy is used as evaluation metric while the best and second-best results for all methods are highlighted in \textbf{\textcolor{red}{red}} and \textbf{\textcolor{blue}{blue}}, respectively.
\\[8pt]
\noindent \textbf{Downstream Tasks.} First, we compare our MP with SOTA methods on CIFAR-100 and FGVC datasets by using ViT-B/16 model pre-trained on ImageNet-21K. From Table~\ref{table 2} we can observe that (1) our MP consistently outperforms the original LP on all downstream tasks by a large margin, achieving 9.24\% gains on average. We owe these performance gains to exploration of the stronger representations for linear classifier inherent in our MP. (2) Compared to full fine-tuning and other parameter-efficient methods, MP achieves comparable or even better performance at much lower training cost (see Figure~\ref{cost}), showing the potential of MP in tuning pre-trained models. (3) By considering the effect of MP on feature learning, MP$_{+}$ achieves further improvement and surpasses all compared tuning methods, verifying the effectiveness of MP on feature learning. 

\noindent \textbf{Backbone Models.} Furthermore, we compare our MP with SOTA methods using four backbone models, including ViT-B/16~\cite{dosovitskiy2021an}, Swin-B~\cite{liu2021swin}, ConvNeXt-B~\cite{liu2022convnet} and AS-MLP~\cite{Lian_2021_ASMLP}, which cover Vision Transformers (ViTs), deep CNNs and MLP-Mixer. Specifically, ViT-B/16, Swin-B and ConvNeXt-B are pre-trained on ImageNet-21K and fine-tuned on ImageNet-1K, while AS-MLP is pre-trained on ImageNet-1K and fine-tuned on CIFAR-100. Table~\ref{table 3} gives the compared results, where MP outperforms LP by more than 1\% and about 5\% on ImageNet-1K and CIFAR-100, respectively. Besides, MP is superior to existing parameter-efficient methods on ImageNet-1K. Since there exists large domain gap between ImageNet-1K and CIFAR-100, feature learning is necessary to achieve promising performance. As such, MP$_{+}$ brings 5.72\% gains over MP for AS-MLP on CIFAR-100. Particularly, our MP$_{+}$ performs better than its counterparts, while obtaining very similar results with full fine-tuning, but it is much more parameter-efficient. These results verify the effectiveness of MP on various backbones.
\begin{table*}[t]
\begin{subtable}[t]{0.355\linewidth}
\scalebox{0.9}{
\centering
\begin{tabular}{c|c|c} 
\specialrule{.1em}{0em}{0em} 
Method & Params.(M) & Acc.(\%) \\ 
\hline
CLS\_token & 0.77 & 82.04 \\
GAP & 0.77 & 79.34 \\
GCP & 4.10 & 80.11 \\
MHC$^{3}$ & 2.88 & 82.10 \\ 
\hline
CLS\_token + GAP & 1.54 & 81.74 \\ 
CLS\_token + GCP & 4.86 & 82.54 \\
\rowcolor{green!20} MP & 3.65 & 83.15 \\
\specialrule{.1em}{0em}{0em} 
\end{tabular}}
\caption{Probing representations.}
\label{ablation a}
\end{subtable}
\begin{subtable}[t]{0.275\linewidth}
\scalebox{0.9}{
\centering
\begin{tabular}{c|c|c} 
\specialrule{.1em}{0em}{0em} 
\multicolumn{1}{l|}{Dimension $\hat{d}$} & \begin{tabular}[c]{@{}c@{}}Params.\\(M)\end{tabular} & \begin{tabular}[c]{@{}c@{}}Acc.\\(\%)\end{tabular} \\ 
\hline
128 & 0.93 & 82.56 \\
256 & 1.44 & 82.72 \\
384 & 2.36 & 82.90 \\
\rowcolor{green!20} 512 & 3.65 & 83.15 \\
640 & 5.44 & 83.24 \\
768 & 7.41 & 83.35 \\
\specialrule{.1em}{0em}{0em} 
\end{tabular}}
\caption{Dimension $\hat{d}$ in MP.}
\label{ablation b}
\end{subtable}
\begin{subtable}[t]{0.39\linewidth}
\scalebox{0.9}{
\begin{tabular}{c|c|c|c|c} 
\specialrule{.1em}{0em}{0em} 
& \multicolumn{2}{c|}{AdaptMLP} & \multicolumn{2}{c}{PSRP} \\ 
\hline
Dim. $d_{h}$ & \begin{tabular}[c]{@{}c@{}}Params.\\(M)\end{tabular} & \begin{tabular}[c]{@{}c@{}}Acc.\\(\%)\end{tabular} & \begin{tabular}[c]{@{}c@{}}Params.\\(M)\end{tabular} & \begin{tabular}[c]{@{}c@{}}Acc.\\(\%)\end{tabular}  \\ 
\hline
4 & 0.84 & 82.62 & 0.88 & 82.80 \\
8 & 0.92 & 82.85 & 0.99 & 83.04 \\
\rowcolor{green!20} 16 & 1.06 & 83.01 & 1.21 & 83.18 \\
32 & 1.36 & 83.08 & 1.65 & 83.31 \\
64 & 1.95 & 83.19 & 2.53 & 83.43 \\
\specialrule{.1em}{0em}{0em} 
\end{tabular}}
\caption{Dimension $d_{h}$ in PSRP.}
\label{ablation c}
\end{subtable}
\caption{Ablation studies on MP and PSRP. Note that the results highlighted \colorbox{green!20}{color} indicate the default settings of our work.}
\label{ablation}
\end{table*}
\\
\noindent\textbf{Comparison of Computational Complexity.} To assess computational efficiency of our MP, we compare with existing methods in terms of training time per batch and GPU memory usage for training, where ViT-B/16~\cite{dosovitskiy2021an} pre-trained on ImageNet-21K is used as a basic backbone. All models are fine-tuned with a batch size of 64 on a single NVIDIA A100 GPU. As shown in  Figure~\ref{cost}, our MP shares similar computational cost with one of LP and it is much more efficient than other existing parameter-efficient methods, because MP and LP avoid gradient computation and update for the intermediate layers. Our  MP$_{+}$ shares similar computational cost with those of parameter-efficient methods, while being more efficient than full fine-tuning and SSF. According to above results, we can conclude that our MP provides a promising solution to achieve a good trade-off between performance and computational complexity, which is more suitable for low-cost computing resource.

\subsection{Ablation Studies}
\label{4.3}
In this subsection, we make ablation studies to evaluate the effect of key components, including probing representations, feature dimension $d$ in MP and hidden dimension $d_{h}$ in PSRP. Besides, we  compare MHC$^3$ with state-of-the-art second-order representations. Here, we use ViT-B/16 pre-trained on ImageNet-21K as backbone and fine-tune it on ImageNet-1K.
\\[5pt]
\noindent \textbf{Comparison of Different Probing Representations.} To assess effect of representations on linear probing, we compare with several representations, including classification token (CLS\_token), global average pooling (GAP) of word tokens, the original global covariance pooling (GCP) of word tokens and their combinations. As shown in Table~\ref{ablation a}, we can see that CLS\_token (weighted combination of word tokens) is superior to GAP. Particularly, the original GCP brings no gain over CLS\_token. In contrast, our MHC$^{3}$ is slightly superior to CLS\_token, while outperforming GCP by $\sim$2\% with fewer parameters. It indicates that our MHC$^{3}$ explores second-order moments for tuning pre-trained models in a more effective and efficient way. By combining CLS\_token with GCP or MHC$^3$, the performance will further increase. However, combination of CLS\_token with GAP leads to inferior performance, which may be caused by both CLS\_token and GAP are first-order statistics, suffering from weak complementary. Our MP explores both first- and second-order moment to achieve the best results, performing better than second-best method (CLS\_token + GCP) by 0.6\% with fewer parameters. Above results clearly demonstrate the effectiveness of representations in our MP for linear probing.
\\[5pt]
\noindent \textbf{Effect of Dimension on MP and PSRP.} For computing our MHC$^{3}$, we first exploit a $1\times1$ convolution layer to reduce dimension of features $\mathbf{X}$ from $d$ to $\hat{d}$ ($\hat{d}<d$). To evaluate effect of dimension $\hat{d}$, we experiment with MP by changing $\hat{d}$ from 128 to 768. As listed in Table~\ref{ablation b}, performance of MP consistently increases when $\hat{d}$ becomes larger. However, larger $\hat{d}$ will involve more parameters. To balance efficiency and effectiveness, we set $\hat{d}$ to 512 throughout all experiments. Furthermore, we assess effect of hidden dimension $d_{h}$ in Eqn.~(\ref{formula 12}) on our PSRP, and compare with AdaptMLP~\cite{chen2022adaptformer}. As compared in Table~\ref{ablation c}, larger $d_{h}$ leads to better performance but more parameters for both AdaptMLP and PSRP. Meanwhile, our PSRP consistently outperforms AdaptMLP with few additional parameters, verifying the effectiveness of PSRP. In our work, $d_{h}$ is set to 16 for efficiency and effectiveness trade-off.
\\[5pt]
\noindent \textbf{Comparison of Second-order Moment} 
To evaluate the efficiency and effectiveness of MHC$^3$ for computing second-order moment, we compare with two state-of-the-art second-order representations (i.e., B-CNN~\cite{lin2015bilinear} and iSQRT-COV~\cite{li2018towards}) on ImageNet-1k, where ViT-B/16 pretrained on ImageNet-21K is used as the backbone. To achieve a good efficiency and effectiveness trade-off, we set feature dimension $\hat{d}$ to 128 for all compared second-order representations. As shown in Table~\ref{more rep}, our MHC$^3$ improves existing second-order representations more than 1.71\% in Top-1 accuracy, while having fewer parameters. Furthermore, our MP is superior to all combinations of  classification token (CLS\_token) with other second-order representations in terms of both efficiency and effectiveness. These results above clearly demonstrate that MHC$^3$ involved in our MP is more suitable for tuning pre-trained models.

\begin{figure*}[ht]
\includegraphics[width=1.0\linewidth,trim=0 15 0 10,clip]{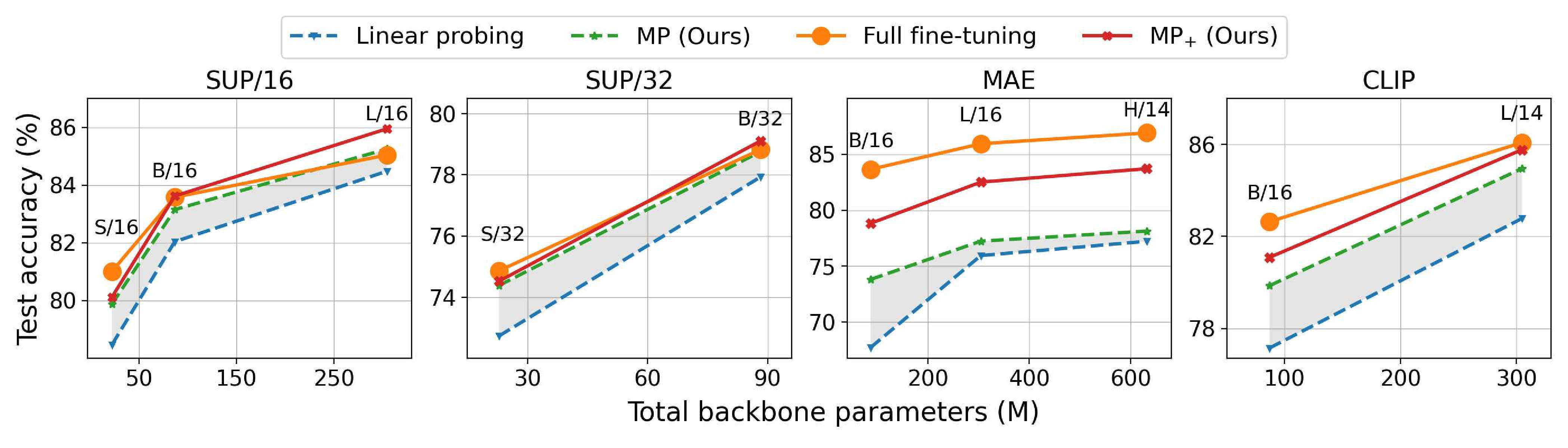}
\caption{Comparisons across different pre-trained objectives and scales. The shadow region indicates performance gap between MP and LP. The size of markers is proportional to number of trainable parameters in log scale.}
\label{sup_mae_clip}
\end{figure*}
\begin{table}[t]
\centering
\begin{tabular}{l|c|c} 
\specialrule{.1em}{0em}{0em} 
Method & Params.(M) & Acc.(\%) \\ 
\hline
GCP & 4.10 & 80.11 \\
B-CNN~\cite{lin2015bilinear} & 4.10  & 80.38\\
iSQRT-COV~\cite{li2018towards} & 4.10 & 80.39\\
MHC$^{3}$ & \textbf{2.88} & \textbf{82.10} \\ 
\hline
CLS\_token + GCP & 4.86 & 82.54 \\
CLS\_token + B-CNN & 4.86  &82.68 \\
CLS\_token + iSQRT-COV & 4.86 & 82.62 \\
\hline
MP (Ours) & \textbf{3.65} & \textbf{83.15} \\
\specialrule{.1em}{0em}{0em} 
\end{tabular}
    \caption{Comparison of second-order representations on IN-1K in terms of tuning parameters (Params.) and Top-1 accuracy (Acc.).}
    \label{more rep}
\end{table}

\subsection{Robustness to OOD Setting}
\label{4.4}
To verify the robustness of our MP, we conduct experiments on three out-of-distribution (OOD) datasets, including ImageNet-A (IN-A)~\cite{hendrycks2021natural}, ImageNet-R (IN-R)~\cite{hendrycks2021many} and ImageNet-C (IN-C)~\cite{hendrycks2019robustness}. Specifically, we first fine-tune ViT-Base/16 model pre-trained on ImageNet-21K by using ImageNet-1K (IN-1K), and directly perform inference on three OOD datasets without any training. Here we compare our MP/MP$_{+}$ with several SOTA methods in Table~\ref{ood}, where we can observe that our MP improves LP over about 3\%$\sim$5\% on three OOD datasets. Meanwhile, MP is very competitive to existing parameter-efficient methods at lower training cost. Particularly, full fine-tuning is high-performance on IN-1K, but it shows weak generalization to OOD datasets. On the contrary, our MP$_{+}$ shows high-performance on both IN-1K and OOD datasets. MP$_{+}$ achieves the best results on IN-A and IN-C, while being comparable to the recently proposed SSF on IN-R. Above results demonstrate our MP and MP$_{+}$ have the ability to equip pre-trained models with stronger robustness to OOD.
\begin{table}[t]
\centering
\scalebox{0.75}{
\begin{tabular}{l|c|c|c|c} 
\specialrule{.1em}{0em}{0em} 
\diagbox{Method}{Dataset} & IN-1K (↑) & IN-A (↑) & IN-R (↑) & IN-C (↓)  \\ 
\hline
Linear probing & 82.04     & 33.91    & 52.87    & 46.91     \\ 
MP~(Ours) & 83.15$_{(1.11)}$     & 39.14$_{(5.23)}$    & 55.91$_{(3.04)}$     & 41.75$_{(5.16)}$   \\
\hline
Adapter~\cite{chen2022adaptformer} & 82.72     & 42.21    & 54.13    & 42.65     \\
Bias~\cite{ben-zaken-etal-2022-bitfit} & 82.74     & 42.12    & 55.94    & 41.90     \\
VPT-Shallow~\cite{jia2022vpt} & 82.08     & 30.93    & 53.72    & 46.88     \\
VPT-Deep~\cite{jia2022vpt}& 82.45     & 39.10    & 53.54    & 43.10     \\
AdaptFormer~\cite{chen2022adaptformer}& 83.01 & 42.96&54.45 &42.35\\
SSF~\cite{Lian_2022_SSF}& 83.10     & \textbf{\textcolor{blue}{45.88}}    & \textbf{\textcolor{blue}{56.77}}   & 41.47    \\ 
\hline
Full fine-tuning & \textbf{\textcolor{blue}{83.58}}    & 34.49    & 51.29    & 46.47     \\
MP + VPT-Deep &83.31 &39.76 &54.21 &41.92\\
MP + AdaptFormer &83.43 &43.59 &54.89 & 41.33\\
MP + SSF &83.56 &45.56 &\textcolor{red}{\textbf{56.95}} &\textbf{\textcolor{blue}{41.32}}\\

MP$_{+}$~(Ours)  &\textbf{\textcolor{red}{83.62}} &\textbf{\textcolor{red}{46.11}} & 56.67 &\textbf{\textcolor{red}{41.14}}\\
\specialrule{.1em}{0em}{0em} 
\end{tabular}}
\caption{Comparison of various fine-tuning methods on out-of-distribution datasets, where Top-1 accuracy (\%) is used as metric on IN-1K, IN-A and IN-R and mean corruption error is used as metric on IN-C. Note that ↑ and ↓ indicate the higher and the lower are better, respectively.}
\label{ood}
\end{table}
\subsection{Generalization to Parameter-efficient Methods}
\label{4.5}
To verify the generalization of our MP, we further evaluate the effect of our MP by combining it with other parameter-efficient methods (ie, VPT~\cite{jia2022vpt}, AdaptFormer~\cite{chen2022adaptformer}, SSF~\cite{Lian_2022_SSF}). Specifically, we keep the experimental settings in Sec.~\ref{4.4}. As shown in Table~\ref{ood}, we can see that MP brings 0.86\%, 0.58\% and 0.46\% improvement gains for VPT-Deep, AdaptFormer, and SSF in IN-1K, respectively. Besides, MP has an ability to improve the robustness of existing parameter-efficient methods to OOD settings. These results above verify the effectiveness and complementarity of our MP to existing parameter-efficient methods. Additionally, our MP$_{+}$ outperforms all combinations of MP with existing parameter-efficient methods, demonstrating the superiority of our partially shared module to learn two recalibrating parameters (PSRP) over existing parameter-efficient methods.

\subsection{Generalization to Pre-training Strategies}
\label{4.6}
To explore the effect of different pre-training strategies, we conduct experiments by using various backbones under supervised (SUP~\cite{dosovitskiy2021an}) and self-supervised (MAE~\cite{he2022masked}, CLIP~\cite{radford2021learning}) settings. For SUP setting, we evaluate our methods using five ViT models (i.e., ViT-Small/16, ViT-Small/32, ViT-Base/16, ViT-Base/32, ViT-Large/16) those are pre-trained on ImageNet-21K, and fine-tune them on ImageNet-1K. As shown in the left two columns of Figure~\ref{sup_mae_clip}, our MP consistently outperforms LP by a large margin. Besides, our MP obtains comparable or even better results with full fine-tuning, when model sizes increase. Particularly, MP (85.25\%, 4.4M) and MP$_{+}$ (85.95\%, 5.6M) outperform full fine-tuning (85.04\%, 304.3M) by 0.21\% and 0.71\% with tuning much fewer parameters for ViT-L/16. For self-supervised objectives (i.e., MAE and CLIP), we use ViT-Base/16, ViT-Large/14, ViT-Large/16 and ViT-Huge/14 as basic backbones. As shown in the right two columns of Figure~\ref{sup_mae_clip}, MP still improves LP by a clear margin, while MP$_{+}$ brings further performance gains over MP. Since MAE aims to optimize parameters for image reconstruction, which suffers from a clear gap for classification task and may require to tune amount of parameters for performing model transfer. As such, full fine-tuning achieves the best results. For CLIP, our MP$_{+}$ achieves comparable results with full fine-tuning especially on large-size model, but it is more parameter-efficient. These results show our MP methods can generalize well to different pre-training strategies.

\begin{figure}[t]
\centering
   \includegraphics[width=1.0\linewidth,trim=0 10 0 10,clip]{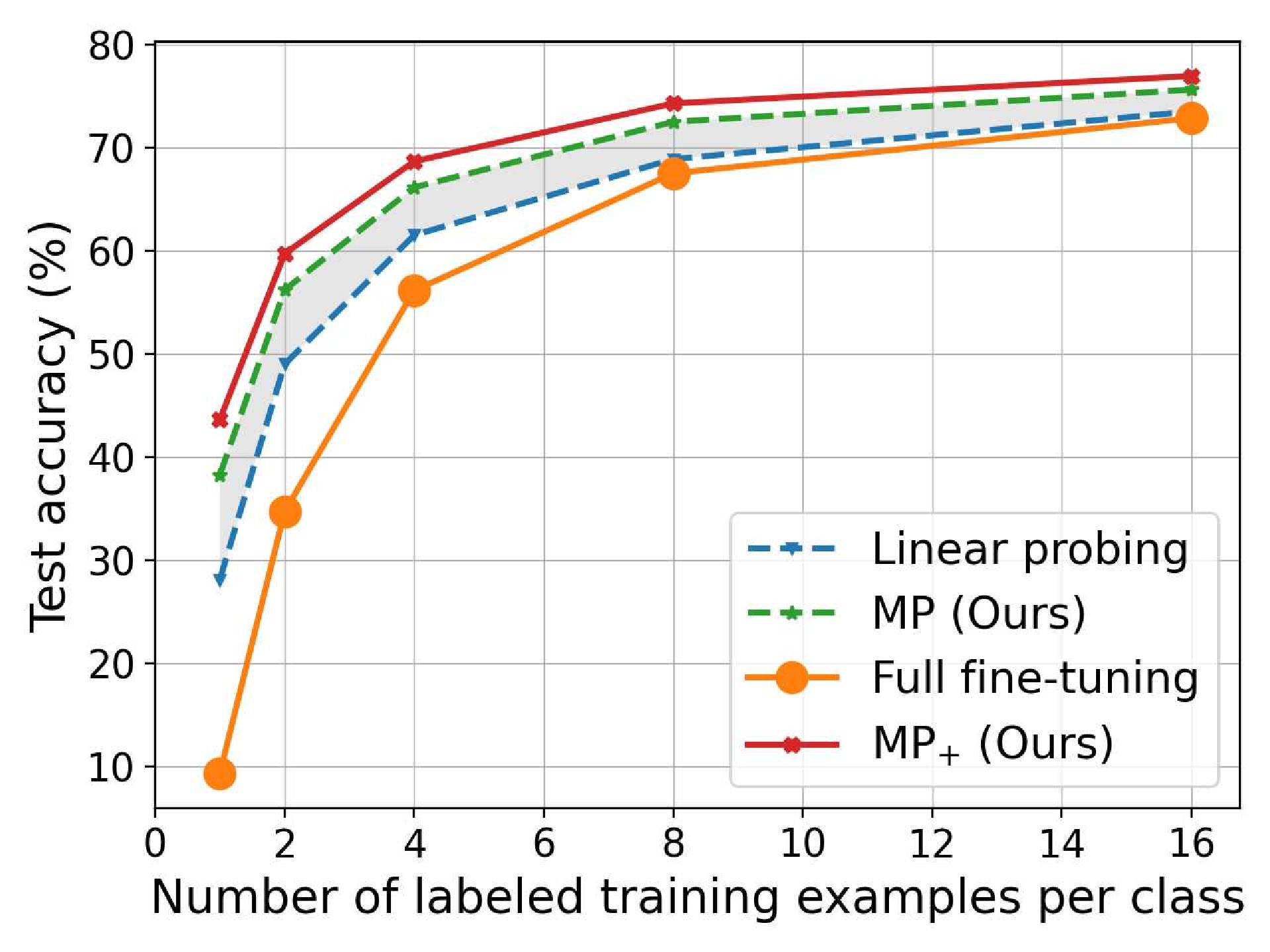}
   \caption{Comparison of various fine-tuning methods under few-shot setting. The shadow region indicates performance gap between MP and LP. The size of markers is proportional to number of trainable parameters in log scale.}
\label{few shot}
\end{figure}

\subsection{Generalization to Few-shot Setting}
\label{4.7}
Finally, we evaluate the performance of our MP and MP$_{+}$ under few-shot setting. Following the setting in~\cite{radford2021learning, zhang2022tip}, we conduct experiments on ImageNet-1K by selecting $\{1, 2, 4, 8, 16\}$ samples for each class as the training set, where ViT-B/16 model pre-trained on ImageNet-21K is used as the backbone. For each setting, we make three trials and report the average results in Figure~\ref{few shot}. It can be seen that full fine-tuning is inferior to LP, which may be caused by that full fine-tuning requires to learn amount of parameters and is difficult to optimize on tiny/small training sets. Our MP is superior to LP for all cases, while MP$_{+}$ achieves further performance improvement by considering parameter-efficient feature learning. The comparisons above demonstrate that parameter-efficient designs for powerful representations and feature learning encourage our MP methods show good generalization to few-shot samples.  

\section{Conclusion}

In this paper, we made an attempt to explore the potential of LP for tuning pre-trained models from the perspective of representations for linear classifiers. Particularly, we propose a Moment Probing (MP) method to feed a powerful representation characterized by feature distribution into classifier, where feature distribution is approximated by combining the original first-order moment of features with an efficient second-order moments (i.e.,  multi-head convolutional cross-covariance, MHC$^3$). Extensive experiments on various settings (e.g., FGVC, different backbones, out-of-distribution, few-shot samples and pre-training strategies) demonstrate the effectiveness and efficiency of MP on tuning pre-trained models. By introducing PSRP modules, our MP$_{+}$ achieves state-of-the-art performance by considering feature learning in a parameter-efficient manner. In the future, we will investigate to extend our MP in prompt learning task~\cite{zhou2022learning}.\\

\noindent \large{\textbf{Acknowledgment}}
\\[5pt]
\normalsize The work was sponsored by National Natural Science Foundation of China (Grant No.s 62276186, 61925602), 
and Haihe Lab of ITAI (NO. 22HHXCJC00002).

{\small
\bibliographystyle{ieee_fullname}
\bibliography{egbib}
}

\end{document}